# Faster Gaussian Summation: Theory and Experiment

**Dongryeol Lee**
College of Computing
Georgia Institute of Technology
Atlanta, GA 30332

**Alexander Gray**
College of Computing
Georgia Institute of Technology
Atlanta, GA 30332

## Abstract

We provide faster algorithms for the problem of Gaussian summation, which occurs in many machine learning methods. We develop two new extensions - an $O(D^p)$ Taylor expansion for the Gaussian kernel with rigorous error bounds and a new error control scheme integrating any arbitrary approximation method - within the best discrete-algorithmic framework using adaptive hierarchical data structures. We rigorously evaluate these techniques empirically in the context of optimal bandwidth selection in kernel density estimation, revealing the strengths and weaknesses of current state-of-the-art approaches for the first time. Our results demonstrate that the new error control scheme yields improved performance, whereas the series expansion approach is only effective in low dimensions (five or less).

## 1 Fast Gaussian Summation

Kernel summations occur ubiquitously in both old and new machine learning algorithms, including kernel density estimation, kernel regression, radial basis function networks, spectral clustering, and kernel PCA (Gray & Moore, 2001; de Freitas et al., 2006). This paper will focus on the most common form $G(x_q) = \sum_{r=1}^{N} w_r K(\delta_{qr})$ in which we desire the sum for $M$ different *query points* $x_q$'s, each using $N$ *reference points* $x_r$'s weighted by $w_r > 0$. $K(\delta_{qr}) = e^{\frac{-\delta_{qr}^2}{2h^2}}$ is the Gaussian kernel, where $\delta_{qr} = ||x_q - x_r||$ with scaling parameter, or *bandwidth* $h$. For concreteness we will take as our main example kernel density estimation (Silverman, 1986), the most widely used distribution-free method for the fundamental task of density estimation. Because the Gaussian kernel has infinite tail, we must pursue approximation in order to achieve runtimes less than that of exhaustive summation. Our goal is to compute each $\widetilde{G}(x_q)$ as quickly as possible while ensuring that $\forall x_q \quad \frac{|\widetilde{G}(x_q)-G(x_q)|}{G(x_q)} \leq \epsilon$ where $\epsilon$ is a user-supplied error tolerance. In practice we wish to perform this computation for a range of bandwidths, from small to large, for example in order to do optimal bandwidth selection by cross-validation.

The basic idea in kernel summation is to approximate the kernel sum contribution $G_R(x_q)$ of some subset of the reference points $X_R$ of size $N_R$, lying in some compact region of space $R$ with centroid $x_R$, to a query point. In more efficient schemes the approximate contribution is made to an entire subset of the query points $X_Q$ of size $N_Q$ lying in some region of space $Q$, with centroid $x_Q$.

**Methods from computational physics.** The successful Fast Multipole Method (FMM) (Greengard & Rokhlin, 1987) developed for the Coulombic kernel, used multipole expansions for the continuous approximation, octtrees (a form of hierarchical grid) for the discrete data structure, and an explicit level-by-level enumeration of the node-node comparisons. Since the expansions only hold locally, (Greengard & Rokhlin, 1987) developed a set of three 'translation operators' for converting between expansions centered at different points in order to create their hierarchical algorithm.

The original Fast Gauss Transform (FGT) (Greengard & Strain, 1991) was developed in the same style, but for the Gaussian kernel using two different expansions. The first one is the multivariate *Hermite expansion* which expresses a sum as an expansion about a representative centroid $x_R$ in the reference region $R$: [1]

---

[1] In this paper we use the multi-index notation (Greengard & Strain, 1991; Yang et al., 2003). A multi-index $\alpha = (\alpha_1, \alpha_2, ..., \alpha_D)$ is a $D$-tuple of integers. For any multi-index $\alpha$, $\beta$ and any $x \in \mathbb{R}^D$, $(1)|\alpha| = \alpha_1 + \alpha_2 + \cdots + \alpha_D, (2)\alpha! = \alpha_1!\alpha_2!\cdots\alpha_D!, (3)x^\alpha = x^{\alpha_1}x^{\alpha_2}\cdots x^{\alpha_D}, (4)D^\alpha = \partial_1^{\alpha_1}\partial_2^{\alpha_2}\cdots\partial_D^{\alpha_D}, (5)\alpha + \beta = (\alpha_1 + $

$$G(x_q) = \sum_{x_r \in R} w_r \sum_{|\alpha| \geq 0} \frac{1}{\alpha!} \left(\frac{x_r - x_R}{\sqrt{2h^2}}\right)^\alpha h_\alpha\left(\frac{x_q - x_R}{\sqrt{2h^2}}\right)$$

This can be re-written as:

$$G(x_q) = \sum_{x_r \in R} w_r \sum_{|\alpha| \geq 0} \frac{1}{\alpha!} h_\alpha\left(\frac{x_r - x_Q}{\sqrt{2h^2}}\right) \left(\frac{x_q - x_Q}{\sqrt{2h^2}}\right)^\alpha$$

as a *Taylor (local) expansion* about a representative centroid $x_Q$ in the query region. [2]

**Dual-tree recursion.** In terms of discrete algorithmic structure, the dual-tree framework of (Gray & Moore, 2001), in the context of kernel summation, generalizes all of the well-known algorithms, including the Barnes-Hut algorithm (Barnes & Hut, 1986), the Fast Multipole Method (Greengard & Rokhlin, 1987), Appel's algorithm (Appel, 1985), and the WSPD (Callahan & Kosaraju, 1995): it is a node-node algorithm (considers query regions rather than points), is fully recursive, can use adaptive data structures such as *kd*-trees, and is bichromatic (can specialize for differing query and reference sets). The idea is to represent both the query points and the reference points respectively with a tree and recurse on a pair of query and reference node. This is shown in depth-first form in Figure 1 though it can also be performed using a priority queue (Gray & Moore, 2003a). It was applied

```
Dualtree(Q, R)
  if Can-approximate(Q, R, ϵ)
    Approximate(Q, R), return
  if leaf(Q) and leaf(R), DualtreeBase(Q, R)
  else
    Dualtree(Q.l, R.l), Dualtree(Q.l, R.r)
    Dualtree(Q.r, R.l), Dualtree(Q.r, R.r)
```

Figure 1: Generic structure of a dual-tree algorithm.

to the problem of kernel density estimation in (Gray & Moore, 2003b) using a finite-difference approximation, a variant of a monopole approximation. Partially by avoiding series expansions, which depend explicitly on the dimension, the result was the fastest such algorithm for general dimension, when operating at the optimal bandwidth. However, when performing cross-validation to determine the (initially unknown) optimal bandwidth, both suboptimally small and large bandwidths must be evaluated. This finite-difference-based method tends to be efficient around or below the optimal bandwidth, and at very large bandwidths, but for intermediately-large bandwidths it suffers.

---

$\beta_1, \cdots, \alpha_D + \beta_D), (6) \alpha - \beta = (\alpha_1 - \beta_1, \cdots, \alpha_D - \beta_D)$, where $\partial_i$ is a $i$-th directional partial derivative. We define $\alpha > \beta$ if $\alpha_i > \beta_i$, and $\alpha \geq p$ for $p \in \mathbb{Z}$ if $\alpha_i \geq p$ for $1 \leq i \leq D$.

[2] We define the Hermite functions $h_n(t)$ by $h_n(t) = e^{-t^2} H_n(t)$, where the Hermite polynomials $H_n(t)$ are defined by the Rodrigues formula: $H_n(t) = (-1)^n e^{t^2} D^n e^{-t^2}$, $t \in \mathbb{R}^1$. The multivariate Hermite function is then defined as a product of its univariate versions: $h_\alpha(t) = \prod_{d=1}^D h_{\alpha_d}(t)$.

**Automatic error control.** Among the existing methods, dual-tree method is the only one to automatically achieve the user's error tolerance $\epsilon$. Other methods are overridden with many *tweak parameters* whose values have to be changed simultaneously with little or no guidance. These parameters waste human time and offer no error tolerance guarantee (unless verified by a procedure that computes density estimate exhaustively). This issue is discussed in Section 7.

**Series expansion.** Expansions in (Greengard & Strain, 1991) require the computation of $O(p^D)$ sub-terms. While effective in the context of computational physics problems, this is problematic in statistical/data mining applications, in which $D$ may be larger than 2 or 3. (Lee et al., 2006) developed the translation operators and error bounds necessary to perform the original FGT-style $O(p^D)$ approximation within the context of the dual-tree framework, demonstrating the first hierarchical fast Gauss transform. However, the new algorithm showed efficiency over any of the aforementioned methods over the entire range of bandwidths necessary in cross-validation, only in very low dimensions (3 or less). The Improved Fast Gauss Transform (IFGT) (Yang et al., 2003) introduced a rearranged series approximation requiring $O(D^p)$ sub-terms, which seemed promising for higher dimensions with an associated error bound, which was unfortunately incorrect. The IFGT was based on a flat set of clusters and did not provide any translation operators.

**This paper.** We demonstrate for the first time the $O(D^p)$ (rather than $O(p^D)$) expansion of the Gaussian kernel (different from that of the IFGT) within a *hierarchical* (dual-tree) algorithm. We also introduce a more efficient mechanism for automatically achieving the user's error tolerance which works with both discrete and continuous approximation schemes. We evaluate these new techniques empirically on real datasets, revealing the strengths and weaknesses of the main current approaches for the first time.

## 2 $O(D^p)$ and $O(p^D)$ Expansions

For concreteness, we first discuss the difference between $O(p^D)$ and $O(D^p)$ expansion by approximating the 2-D Gaussian kernel using its *Hermite expansion* at order $p = 2$. Its $O(p^D)$ expansion is:

$$e^{\frac{-||x_q - x_r||^2}{2h^2}}$$

$$= \prod_{d=1}^{2} \left( h_0\left(\frac{x_{qd} - x_{Rd}}{\sqrt{2h^2}}\right) + \left(\frac{x_{rd} - x_{Rd}}{\sqrt{2h^2}}\right) h_1\left(\frac{x_{qd} - x_{Rd}}{\sqrt{2h^2}}\right) + \cdots \right)$$

$$\approx 1 \cdot h_0\left(\frac{x_{q1} - x_{R1}}{\sqrt{2h^2}}\right) h_0\left(\frac{x_{q2} - x_{R2}}{\sqrt{2h^2}}\right) + \left(\frac{x_{r2} - x_{R2}}{\sqrt{2h^2}}\right) h_0\left(\frac{x_{q1} - x_{R1}}{\sqrt{2h^2}}\right)$$
$$h_1\left(\frac{x_{q2} - x_{R2}}{\sqrt{2h^2}}\right) + \left(\frac{x_{r1} - x_{R1}}{\sqrt{2h^2}}\right) h_1\left(\frac{x_{q1} - x_{R1}}{\sqrt{2h^2}}\right) h_0\left(\frac{x_{q2} - x_{R2}}{\sqrt{2h^2}}\right) +$$
$$\left(\frac{x_{r1} - x_{R1}}{\sqrt{2h^2}}\right) \left(\frac{x_{r2} - x_{R2}}{\sqrt{2h^2}}\right) h_1\left(\frac{x_{q1} - x_{R1}}{\sqrt{2h^2}}\right) h_1\left(\frac{x_{q2} - x_{R2}}{\sqrt{2h^2}}\right)$$

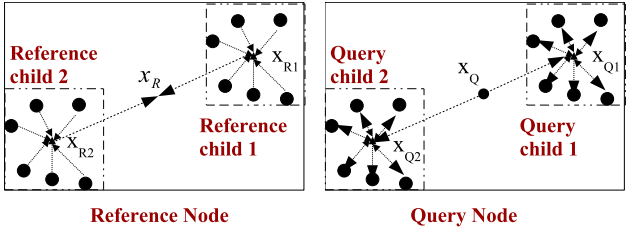

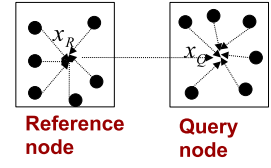

Figure 2: Left: *H2H operator* combines two finer scaled *far field expansions* centered at $x_{R1}$ and $x_{R2}$ into a coarser scaled one centered $x_R$. Right: *L2L operator* converts the *local expansion* centered at $x_Q$ into two finer scaled ones centered at $x_{Q1}$ and $x_{Q2}$.

Figure 3: *H2L operator* converts the *far field expansion* centered at $x_R$ into the *local expansion* centered at $x_Q$.

On the other hand, $O(D^p)$ expansion uses *graded lexicographic ordering* (Yang et al., 2003), and yields:

$$e^{\frac{-\|x_q - x_r\|^2}{2h^2}}$$
$$\approx 1 \cdot h_0\left(\tfrac{x_{q1}-x_{R1}}{\sqrt{2h^2}}\right)h_0\left(\tfrac{x_{q2}-x_{R2}}{\sqrt{2h^2}}\right) + \left(\tfrac{x_{r1}-x_{R1}}{\sqrt{2h^2}}\right)h_1\left(\tfrac{x_{q1}-x_{R1}}{\sqrt{2h^2}}\right)$$
$$h_0\left(\tfrac{x_{q2}-x_{R2}}{\sqrt{2h^2}}\right) + \left(\tfrac{x_{r2}-x_{R2}}{\sqrt{2h^2}}\right)h_0\left(\tfrac{x_{q1}-x_{R1}}{\sqrt{2h^2}}\right)h_1\left(\tfrac{x_{q2}-x_{R2}}{\sqrt{2h^2}}\right)$$

In both cases, $\prod_{d=1}^{D} h_{\alpha_d}\left(\frac{x_{qd}-x_{Rd}}{\sqrt{2h^2}}\right)$ forms basis functions for the expansion. In actual codes, the factors in front of these basis functions are stored as coefficients in the corresponding *reference node*. $O(p^D)$ expansion requires exactly $p^D$ coefficients, while $O(D^p)$ one requires $\binom{D+p-1}{D}$. If we iterate over all *reference points* in the *reference node* with their weights taken into account, we store: $\sum_{r=1}^{N_R} w_r$, $\sum_{r=1}^{N_R} w_r\left(\frac{x_{r2}-x_{R2}}{\sqrt{2h^2}}\right)$, $\sum_{r=1}^{N_R} w_r\left(\frac{x_{r1}-x_{R1}}{\sqrt{2h^2}}\right)$, $\sum_{r=1}^{N_R} w_r\left(\frac{x_{r1}-x_{R1}}{\sqrt{2h^2}}\right)\left(\frac{x_{r2}-x_{R2}}{\sqrt{2h^2}}\right)$ and $\sum_{r=1}^{N_R} w_r$, $\sum_{r=1}^{N_R} w_r\left(\frac{x_{r1}-x_{R1}}{\sqrt{2h^2}}\right)$, $\sum_{r=1}^{N_R} w_r\left(\frac{x_{r2}-x_{R2}}{\sqrt{2h^2}}\right)$ for $O(p^D)$ and $O(D^p)$ expansions respectively. The *Taylor expansion* works similarly, except that the coefficients are stored in the corresponding *query node*.

## 3 Translation Operators

Since the properties of the Gaussian kernel do not require that approximation be made in the local fashion, the original FGT used a flat grid with only **H2L** operator whose associated incorrect error was corrected by (Baxter & Roussos, 2002). (Lee et al., 2006) derived two additional translation operators necessary for a *hierarchical* FGT and the associated error bounds for $O(p^D)$ expansion of Hermite/Taylor coefficients. We briefly review all three translation operators.

The first translation operator transfers the contribution of a reference node $R$ into the Taylor series centered about $x_Q$ in a query node $Q$.

**Lemma 1.** Hermite-to-local (H2L) translation operator *(in Lemma 2.2 in (Greengard & Strain, 1991)): Given a reference node $R$, a query node $Q$, and the Hermite expansion centered at a centroid $x_R$ of $R$: $G(x_q) = \sum_{|\alpha| \geq 0} A_\alpha h_\alpha\left(\frac{x_q-x_R}{\sqrt{2h^2}}\right)$ where $A_\alpha = \sum_{r=1}^{N_R} \frac{w_r}{\alpha!}\left(\frac{x_r-x_R}{\sqrt{2h^2}}\right)^\alpha$, the Taylor expansion at the centroid $x_Q$ of $Q$ is given by: $G(x_q) = \sum_{|\beta| \geq 0} B_\beta \left(\frac{x_q-x_Q}{\sqrt{2h^2}}\right)^\beta$ where $B_\beta = \frac{(-1)^\beta}{\beta!} \sum_{|\alpha| \geq 0} A_\alpha h_{\alpha+\beta}\left(\frac{x_Q-x_R}{\sqrt{2h^2}}\right)$.*

The next operator allows efficient precomputation of the Hermite moments in the reference tree in a bottom-up fashion from its children.

**Lemma 2.** Hermite-to-Hermite (H2H) translation operator: *Given the Hermite expansion centered at a centroid $x_{R'}$ in a reference node $R'$: $G(x_q) = \sum_{|\alpha| \geq 0} A'_\alpha h_\alpha\left(\frac{x_q-x_{R'}}{\sqrt{2h^2}}\right)$ this same Hermite expansion shifted to a new location $x_R$ of the parent node $R$ is given by: $G(x_q) = \sum_{|\gamma| \geq 0} A_\gamma h_\gamma\left(\frac{x_q-x_R}{\sqrt{2h^2}}\right)$ where $A_\gamma = \sum_{0 \leq \alpha \leq \gamma} \frac{1}{(\gamma-\alpha)!} A'_\alpha \left(\frac{x_{R'}-x_R}{\sqrt{2h^2}}\right)^{\gamma-\alpha}$.*

The final operator combines the approximations at different scales through one breadth-first traversal.

**Lemma 3.** Local-to-local (L2L) translation operator: *Given a Taylor expansion centered at a centroid $x_{Q'}$ of a query node $Q'$: $G(x_q) = \sum_{|\beta| \geq 0} B_\beta \left(\frac{x_q-x_{Q'}}{\sqrt{2h^2}}\right)^\beta$ the Taylor expansion obtained by shifting this expansion to the new centroid $x_Q$ of the child node $Q$ is: $G(x_q) = \sum_{|\alpha| \geq 0} \left[\sum_{\beta \geq \alpha} \frac{\beta!}{\alpha!(\beta-\alpha)!} B_\beta \left(\frac{x_Q-x_{Q'}}{\sqrt{2h^2}}\right)^{\beta-\alpha}\right] \left(\frac{x_q-x_Q}{\sqrt{2h^2}}\right)^\alpha$.*

## 4 Error Bounds for $O(D^p)$ Expansions

Because Hermite/Taylor expansions are truncated after a finite number of terms, we incur an error in approximation. In order to bound the total approximation error, we need one error bound for each translation operator. In (Lee et al., 2006), the Hermite and the Taylor expansion were treated as products of $D$ univariate Hermite/Taylor expansions. The trailing sum in each univariate expansion was bounded using the property of infinite geometric series, which in turn limited the size of the query/reference node for pruning to be valid. Here, we use the same tranlsation operators, but instead view each expansion as a *vector* function and use the $O(D^p)$ expansion advocated in (Yang et al., 2003). The new error bounds based on this new expansion scheme depend on the multidimensional Taylor's Theorem, and effectively eliminate the node size restriction imposed by the $O(p^D)$ expansion (Greengard & Strain, 1991; Lee et al., 2006).

**Theorem 1.** Multidimensional Taylor's Theorem: Let $O \subset \mathbb{R}^D$ be an open set. Let $x_* \in O$ and $f$ be a function which is $n$ times differentiable in $O$. For any $x \in O$, there exists $\theta \in \mathbb{R}$ with $0 < \theta < 1$ such that $f(x) = \sum_{|\alpha|<p} \frac{1}{\alpha!} D^\alpha f(x_*)(x-x_*)^\alpha + \sum_{|\alpha|=p} \frac{1}{\alpha!} D^\alpha f(x_* + \theta(x-x_*))(x-x_*)^\alpha$. The last term $R_n = \sum_{|\alpha|=p} \frac{1}{\alpha!} D^\alpha f(x_* + \theta(x-x_*))(x-x_*)^\alpha$ is called the Lagrange remainder and $|R_n| \leq \sum_{|\alpha|=p} \frac{1}{\alpha!} \sup_{0<\theta<1} \left| D^\alpha f(x_* + \theta(x-x_*)) \right| \prod_{d=1}^{D} |x_d - x_{*d}|^{\alpha_d}$

The first lemma gives an upper bound on the absolute error on estimating a reference node contribution by evaluating a truncated Hermite expansion. The second lemma gives an upper bound on the absolute error incurred from approximating the contribution of a reference node by evaluating the Taylor series formed via direct local accumulation of each reference point.

**Lemma 4.** Given a query node $Q$, a reference node $R$ with an Hermite expansion about its centroid $x_R$: $G(x_q) = \sum_{|\alpha|\geq 0} A_\alpha h_\alpha\left(\frac{x_q-x_R}{\sqrt{2h^2}}\right)$, and $x_q \in Q$, the absolute truncating error after taking the first $O(D^p)$ terms is bounded by: $E_{DH}(p) = W_R \frac{e^{\frac{-\delta_{QR}^{min^2}}{4h^2}} \binom{D+p-1}{D-1} r_R^p}{\sqrt{\left(\lfloor \frac{p}{D} \rfloor!\right)^{D-p'} \left(\lceil \frac{p}{D} \rceil!\right)^{p'}}}$ where $r_R = \max_{x_r \in R} \frac{||x_r - x_R||_\infty}{h}$ and $p' = p \mod D$.

*Proof.* By Theorem 1 and the triangle inequality,
$\left| G(x_q) - \sum_{|\alpha|<p} A_\alpha h_\alpha\left(\frac{x_q-x_R}{\sqrt{2h^2}}\right) \right|$
$\leq \sum_{x_r \in R} w_r \left| K(\delta_{qr}) - \sum_{|\alpha|<p} \frac{1}{\alpha!} h_\alpha\left(\frac{x_q-x_R}{\sqrt{2h^2}}\right) \left(\frac{x_r-x_R}{\sqrt{2h^2}}\right)^\alpha \right|$
$\leq W_R \sum_{|\alpha|=p} \frac{1}{\alpha!} \max_{x_q \in Q, x_r \in R} \left| h_\alpha\left(\frac{x_q-x_r}{\sqrt{2h^2}}\right) \right| \prod_{d=1}^{D} \left|\frac{x_{rd}-x_{Rd}}{\sqrt{2h^2}}\right|^{\alpha_d}$
$\leq W_R e^{\frac{-\delta_{QR}^{min^2}}{4h^2}} \sum_{|\alpha|=p} \frac{1}{\alpha!} (\sqrt{2})^\alpha \sqrt{\alpha!} \prod_{d=1}^{D} \left|\frac{x_{rd}-x_{Rd}}{\sqrt{2h^2}}\right|^{\alpha_d}$
$\leq W_R e^{\frac{-\delta_{QR}^{min^2}}{4h^2}} \sum_{|\alpha|=p} \frac{1}{\sqrt{\alpha!}} \prod_{d=1}^{D} \left|\frac{x_{rd}-x_{Rd}}{h}\right|^{\alpha_d}$
$\leq W_R e^{\frac{-\delta_{QR}^{min^2}}{4h^2}} \sum_{|\alpha|=p} \frac{r_R^\alpha}{\sqrt{\alpha!}} \leq W_R \frac{e^{\frac{-\delta_{QR}^{min^2}}{4h^2}} \binom{D+p-1}{D-1} r_R^p}{\sqrt{\left(\lfloor \frac{p}{D} \rfloor!\right)^{D-p'} \left(\lceil \frac{p}{D} \rceil!\right)^{p'}}}$ □

**Lemma 5.** Given the following Taylor expansion about the centroid $x_Q$ of a query node $Q$: $G(x_q) = \sum_{|\beta|\geq 0} B_\beta \left(\frac{x_q-x_Q}{\sqrt{2h^2}}\right)^\beta$ where $B_\beta = \frac{(-1)^{|\beta|}}{\beta!} \sum_{|\alpha|\geq 0} A_\alpha h_{\alpha+\beta}\left(\frac{x_Q-x_R}{\sqrt{2h^2}}\right)$, for any $x_q \in Q$, the absolute error due to truncating the series after $O(D^p)$ terms is bounded by: $E_{DL}(p) = W_R \frac{e^{\frac{-\delta_{QR}^{min^2}}{4h^2}} \binom{D+p-1}{D-1} r_Q^p}{\sqrt{\left(\lfloor \frac{p}{D} \rfloor!\right)^{D-p'} \left(\lceil \frac{p}{D} \rceil!\right)^{p'}}}$ where $r_Q = \max_{x_q \in Q} \frac{||x_q - x_Q||_\infty}{h}$ and $p' = p \mod D$.

*Proof.* The derivation is similar to one in Lemma 4. □

The final lemma gives an upper bound on the absolute error incurred by approximating the reference node contribution by the Taylor expansion converted from the truncated Hermite expansion.

**Lemma 6.** A truncated Hermite expansion about the centroid $x_R$ of a reference node $R$ given by: $G(x_q) = \sum_{|\alpha|<p} A_\alpha h_\alpha\left(\frac{x_q-x_R}{\sqrt{2h^2}}\right)$ has the following Taylor expansion about the centroid $x_Q$ of a query node $Q$: $G(x_q) = \sum_{|\beta|\geq 0} C_\beta \left(\frac{x_q-x_Q}{\sqrt{2h^2}}\right)^\beta$ where $C_\beta = \frac{(-1)^{|\beta|}}{\beta!} \sum_{\alpha<p} A_\alpha h_{\alpha+\beta}\left(\frac{x_Q-x_R}{\sqrt{2h^2}}\right)$. The absolute truncation error after taking $O(D^p)$ terms is: $E_{H2L}(p)$
$= W_R \frac{e^{\frac{-\delta_{QR}^{min^2}}{4h^2}} \binom{D+p-1}{D-1}}{\sqrt{\left(\lfloor \frac{p}{D} \rfloor!\right)^{D-p'} \left(\lceil \frac{p}{D} \rceil!\right)^{p'}}} \left(r_Q^p + \left(\sqrt{2} r_R\right)^p \binom{D+p-1}{D}\right)$
$\left(\sqrt{2} r_Q\right)^{I(\sqrt{2} r_Q)}$ where $r_Q = \max_{x_q \in Q} \frac{||x_q - x_Q||_\infty}{h}$,
$r_R = \max_{x_r \in R} \frac{||x_r - x_R||_\infty}{h}$, $p' = p \mod D$ and
$I(x) = \begin{cases} 0, & 0 \leq x \leq 1 \\ p-1, & otherwise \end{cases}$.

*Proof.* Let $E_1 = \sum_{|\beta|<p} \frac{(-1)^{|\beta|}}{\beta!} \sum_{|\alpha|\geq p} \frac{1}{\alpha!} \left(\frac{x_r-x_R}{\sqrt{2h^2}}\right)^\alpha h_{\alpha+\beta}\left(\frac{x_Q-x_R}{\sqrt{2h^2}}\right) \left(\frac{x_q-x_Q}{\sqrt{2h^2}}\right)^\beta$ and $E_2 = \sum_{|\beta|\geq p} \frac{(-1)^{|\beta|}}{\beta!} h_\beta\left(\frac{x_Q-x_r}{\sqrt{2h^2}}\right) \left(\frac{x_q-x_Q}{\sqrt{2h^2}}\right)^\beta$
Then,
$\left| G(x_q) - \sum_{|\beta|<p} \frac{(-1)^{|\beta|}}{\beta!} \sum_{|\alpha|<p} A_\alpha h_{\alpha+\beta}\left(\frac{x_Q-x_R}{\sqrt{2h^2}}\right) \right|$
$\leq W_R(|E_1| + |E_2|)$

Clearly, $|E_2| \leq \frac{e^{-\frac{\delta_{QR}^{min^2}}{4h^2}} \binom{D+p-1}{D-1} r_Q^p}{\sqrt{\left(\lfloor \frac{p}{D} \rfloor!\right)^{D-p'} \left(\lceil \frac{p}{D} \rceil!\right)^{p'}}}$. In addition,

$|E_1| \leq \sum_{|\beta|<p} \frac{1}{\beta!} \sum_{|\alpha|=p} \frac{1}{\alpha!} \prod_{d=1}^{D} \left|\frac{x_{rd}-x_{Rd}}{\sqrt{2h^2}}\right|^{\alpha_d} \left|\frac{x_{qd}-x_{Qd}}{\sqrt{2h^2}}\right|^{\beta_d}$
$\max_{x_q \in Q, x_r \in R} \left|h_{\alpha+\beta}\left(\frac{x_q-x_r}{\sqrt{2h^2}}\right)\right|$

$\leq e^{-\frac{\delta_{QR}^{min^2}}{4h^2}} \sum_{|\beta|<p} \frac{1}{\sqrt{\beta!}} \sum_{|\alpha|=p} \sqrt{\frac{(\alpha+\beta)!}{\alpha!\beta!}} \frac{\sqrt{2}^{\alpha+\beta}}{\sqrt{\alpha!}}$
$\prod_{d=1}^{D} \left|\frac{x_{rd}-x_{Rd}}{\sqrt{2h^2}}\right|^{\alpha_d} \left|\frac{x_{qd}-x_{Qd}}{\sqrt{2h^2}}\right|^{\beta_d}$

$\leq e^{-\frac{\delta_{QR}^{min^2}}{4h^2}} \sum_{|\beta|<p} \frac{1}{\sqrt{\beta!}} \sum_{|\alpha|=p} \sqrt{2}^{\alpha+\beta} \frac{\sqrt{2}^{\alpha+\beta}}{\sqrt{\alpha!}}$
$\prod_{d=1}^{D} \left|\frac{x_{rd}-x_{Rd}}{\sqrt{2h^2}}\right|^{\alpha_d} \left|\frac{x_{qd}-x_{Qd}}{\sqrt{2h^2}}\right|^{\beta_d}$

$\leq e^{-\frac{\delta_{QR}^{min^2}}{4h^2}} \sum_{|\beta|<p} \frac{1}{\sqrt{\beta!}} \sum_{|\alpha|=p} \frac{1}{\sqrt{\alpha!}} \prod_{d=1}^{D} \left|\frac{\sqrt{2}(x_{rd}-x_{Rd})}{h}\right|^{\alpha_d}$
$\left|\frac{\sqrt{2}(x_{qd}-x_{Qd})}{h}\right|^{\beta_d}$

$\leq \frac{e^{-\frac{\delta_{QR}^{min^2}}{4h^2}} \binom{D+p-1}{D-1} (\sqrt{2}r_R)^p}{\sqrt{\left(\lfloor \frac{p}{D} \rfloor!\right)^{D-p'} \left(\lceil \frac{p}{D} \rceil!\right)^{p'}}} \sum_{|\beta|<p} \frac{1}{\sqrt{\beta!}} \prod_{d=1}^{D} \left|\frac{\sqrt{2}(x_{qd}-x_{Qd})}{h}\right|^{\beta_d}$

$\leq \frac{e^{-\frac{\delta_{QR}^{min^2}}{4h^2}} \binom{D+p-1}{D-1}}{\sqrt{\left(\lfloor \frac{p}{D} \rfloor!\right)^{D-p'} \left(\lceil \frac{p}{D} \rceil!\right)^{p'}}} (\sqrt{2}r_R)^p \binom{D+p-1}{D} (\sqrt{2}r_Q)^{I(\sqrt{2}r_Q)}$ □

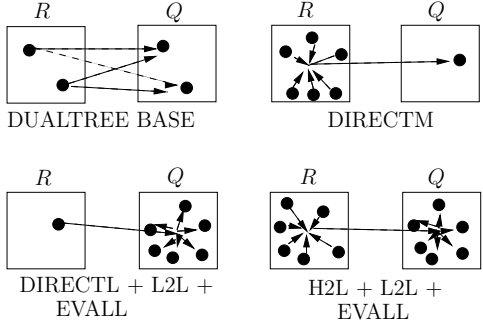

Figure 4: Four ways a reference node can send its contributions to a query node using the original FGT style pruning. In the clockwise order starting from the top left, exhaustive computation (few reference/query points), multipole evaluation (many reference/few query points), direct Taylor accumulation (few reference/many query points), H2L-translation (many reference/query points).

## 5 New Error Guarantee Rule

Let us first revisit our method of automatically guaranteeing the user's error tolerance $\epsilon$ defined in Section 1. We now specify the function **Can-approximate**($Q$,$R$,$\epsilon$), which only has local information (contained in the query node $Q$ and the reference node $R$) available to it, but must guarantee a global error criterion.

In the dual-tree finite-difference algorithm (DFD) (Gray & Moore, 2003b), the function **Approximate**($Q$,$R$) approximates the contribution of $R$ to each query point $x_q$ in $Q$, $G_R(x_q)$, by $\widetilde{G}_R(x_q) = W_R \bar{K} = W_R \frac{K(\delta_{QR}^{max})+K(\delta_{QR}^{min})}{2}$, where $W_R = \sum_{x_r \in R} w_r$ and $\delta_{QR}^{min}$ and $\delta_{QR}^{max}$ are lower and upper bounds on the distance between $x_q \in Q$ and $x_r \in R$, respectively. These distances are easily obtained using the bounding boxes of the nodes. By using these bounds DFD algorithm maintains a running lower bound $G_Q^{min}$ on $G_R(x_q)$ which holds for all $x_q \in Q$. In Section 4, we laid out more approximation methods in addition to finite-difference approximation (**FD**): evaluating a truncated Hermite expansion centered at $x_R$ (**DH**), forming a truncated Taylor expansion centered at $x_Q$ using each reference point (**DL**), and forming an approximated truncated Taylor expansion centered at $x_Q$ by converting the truncated Hermite expansion centered at $x_R$ (**H2L**). The following specifies **Can-approximate**($Q$,$R$,$\epsilon$), which incorporates the new approximation methods.

**Theorem 2.** *Given the following methods for approximating the contribution of a reference node $R$: $\mathbb{A} = \{EX, DH, DL, H2L, FD\}$ where DH, DL, H2L, and FD are denoted as above, and EX for exhaustive computation, $A \in \mathbb{A}$ with a maximum absolute error of $E_A$ can be used to guarantee the global error tolerance $\epsilon$ if: $E_A \leq \frac{W_R \epsilon}{W} G_Q^{min}$ where $W = \sum_{r=1}^{N} w_r$.*

*Proof.* Given $x_q \in Q$, suppose $\widetilde{G}(x_q)$ was computed using $k$ reference nodes $R_i$'s, whose contribution was approximated using $A_i$. By the triangle inequality:

$|G(x_q) - \widetilde{G}(x_q)|$
$= \left|\sum_{i=1}^{k} G_{Ri}(x_q) - \widetilde{G}_{Ri}(x_q)\right| \leq \sum_{i=1}^{k} |G_{Ri}(x_q) - \widetilde{G}_{Ri}(x_q)|$
$\leq \sum_{i=1}^{k} E_{Ai} \leq \sum_{i=1}^{k} \frac{W_{Ri} \epsilon}{W} G_Q^{min} \leq \epsilon G_Q^{min} \leq \epsilon G(x_q)$ □

This new rule generalizes the previous *local* approximation condition (Gray & Moore, 2003b): $|K(\delta_{QR}^{min}) - K(\delta_{QR}^{max})|/G_Q^{min} \leq \frac{2\epsilon}{W}$ where $E_{FD} = \frac{W_R(K(\delta_{QR}^{min})-K(\delta_{QR}^{max}))}{2}$. Clearly, $E_{EX} = 0$, and $E_{DH}$, $E_{DL}$, and $E_{H2L}$ are given as Lemma 4, 5, 6 respectively. The approximation rule above essentially gives each reference node $R$ a maximum relative error proportional to the sum of the weights of reference points it contains. In considering the $i$-th reference node contribution, when $A_i = EX$, the maximum allowable

relative error of $\frac{W_{Ri}\epsilon}{N}$ is not used up; Otherwise, if $G_Q^{min} > 0$, pruning requires only a relative error of $\frac{W'_{Ri}\epsilon}{W}$ where $W'_{Ri} = \frac{WE_{Ai}}{\epsilon G_Q^{min}}$. Our new approximation rule notes that the portion of the weights not used to cover the incurred pruning error can be stored into a field $W_T$ (initialized to zero before the computation and denoted $Q.W_T$ hereon) in each query node $Q$ to use them in future pruning opportunities. The first case yields $W_{Ri}$ as the leftover, while the second case (pruned case) yields $W_{Ri} - W'_{Ri}$.

Given $A \in \mathbb{A}$ with the maximum absolute error of $E_A$, we now modify the approximation condition to: $E_A/G_Q^{min} \leq \frac{\epsilon(W_R + W_T)}{W}$. Solving for $W_T$ yields: $W_T \geq W_R\left(\frac{WE_A}{\epsilon G_Q^{min}} - 1\right)$. Whenever a pruning is attempted, the modified algorithm will evaluate the right handside of the inequality. If the evaluated value is negative, it represents the leftover "token" after pruning is performed and $Q.W_T$ of the current query node will be incremented by $W_R\left(1 - \frac{WE_A}{\epsilon G_Q^{min}}\right)$. If positive, it represents the required extra "token" from the $Q.W_T$ slot of the current query node, in order to prune the given query and reference node pair. If $Q.W_T \geq W_T$, pruning succeeds and $Q.W_T$ is decremented by $W_R\left(\frac{WE_A}{\epsilon G_Q^{min}} - 1\right)$.

## 6 New Dual-tree Algorithm

We first introduce an extra field in each query node $G_Q^{est}$ storing contributions from reference nodes obtained by finite-difference approximation and direct Hermite evaluations. The contributions from Taylor coefficients obtained via direct local accumulation and **H2L** translation opeator will be accounted for during the post-processing step.

In preprocessing, we construct two trees, one for the query dataset and one for the reference dataset. In this paper an efficient form of sphere-rectangle trees (Katayama & Satoh, 1997) is used, with idea of cached sufficient statistics as in mr*kd*-trees (Deng & Moore, 1995). The Hermite moments of order *PLIMIT* is pre-computed for the reference tree. For the experimental results, we have fixed *PLIMIT* = 8 for $D = 2$, *PLIMIT* = 6 for $D = 3$, *PLIMIT* = 4 for $D = 5$, *PLIMIT* = 2 for $D = 6$. We presume that *PLIMIT* = 1 for $D > 6$.

During the recursive function call **DITO**, an optimized version of finite-difference pruning is first attempted. In case of failure, we attempt FMM-type pruning in which we choose the cheapest operation given a query node $Q$ and a reference node $R$ from the followings: direct Hermite evaluation (**DIRECTM***(Hermite coefficients, truncation order, query point)*), direct local accumulation (**DI-**

```
buildInternal(R)
  n = empty node
  {R_1, R_2} = Partition R into two
  n.l = buildReferenceTree(R_1)
  n.r = buildReferenceTree(R_2)
  n.mcoeffs = H2H(n.l.mcoeffs, n.l.x_R, n.x_R)
  n.mcoeffs+ = H2H(n.r.mcoeffs, n.r.x_R, n.x_R)
  return n
buildLeaf(R)
  n = empty node
  n.mcoeffs = Compute the Hermite series of order
    PLIMIT using each x_r ∈ R centered at x_R
  return n
buildReferenceTree(R)
  if size(R) < leaf threshold, return buildLeaf(R)
  else, return buildInternal(R)
```

Figure 5: Building the tree from the *reference dataset*.

**RECTL***(the set of reference points, truncation order, query node centroid)*), H2L translation (**H2L***(Hermite coefficients, truncation order, reference node centroid, query node centroid)*), and exhaustive computations. Roughly, direct Hermite evaluations at each $x_q \in Q$ is $O(N_Q D^{p_{DH}+1})$, direct local accumulation $O(N_R D^{p_{DL}+1})$, H2L translation $O(D^{2p_{H2L}+1})$, an exhaustive method $O(DN_Q N_R)$. In our algorithm, if an exhaustive method is selected, we let the recursion continue, hoping pruning can occur in the finer level of recursion. It is possible to hand-tune the exact cutoffs for determining the optimal choice, but these rough approximations seem to work well.

In the post-processing step, we perform a breadth-first traversal of the query tree. For an internal node $Q$, its Taylor expansion is shifted to the centers of its children via L2L translation operator (**L2L***(Taylor coefficients, truncation order, old query node centroid, new query node centroid)*); the estimated contributions $G_Q^{est}$ are propagated downward directly to its children as well. For a leaf node $Q$, we evaluate the Taylor expansion at every $x_q \in Q$ using **EVALL***(Taylor coefficients, truncation order, query node centroid, query point)* and add in the far-field contribution $G_Q^{est}$.

## 7 Experiments and Conclusions

We empirically evaluated the runtime performance of six algorithms on six real-world datasets (astronomy (2-D), physical simulation (3-D), pharmaceutical (5-D), biology (7-D), forestry (10-D), image textures (16-D)) scaled to fit in $[0, 1]^D$ hypercube, for kernel density estimation at every query point with a range of bandwidths, from 3 orders of magnitude smaller than optimal to three orders larger than optimal, according to the standard least-squares cross-validation scores (Silverman, 1986). In our case, the set of reference points

```
bestMethod(Q, R, maxerr)
  p_DH = the smallest 1 ≤ p ≤ PLIMIT such that
    E_DH(p) ≤ maxerr, otherwise p_DH = ∞
  p_DL = the smallest 1 ≤ p ≤ PLIMIT such that
    E_DL(p) ≤ maxerr, otherwise p_DL = ∞
  p_H2L = the smallest 1 ≤ p ≤ PLIMIT such that
    E_H2L(p) ≤ maxerr, otherwise p_H2L = ∞
  c_DH = N_Q D^{p_DH+1}, c_DL = N_R D^{p_DL+1}
  c_H2L = D^{2p_H2L+1}, c_Direct = D N_Q N_R
  c = min(c_DH, c_DL, c_H2L, c_Direct)
  if c = c_DH, return {DH, p_DH, ε_DH(p_DH)}
  else if c = c_DL, return {DL, p_DL, ε_DL(p_DL)}
  else if c = c_H2L, return {H2L, p_H2L, ε_H2L(p_H2L)}
  else, return {DIRECT, ∞, ∞}
```

Figure 6: Choosing the FMM-type approximation with the least cost for a query and reference node pair.

is the same as the set of query points. All datasets have 50K points so that the exact exhaustive method can be tractably computed. We set the tolerance $\epsilon = 0.01$. We compare: **FGT** (Fast Gauss Transform (Greengard & Strain, 1991)), **IFGT** (Improved Fast Gauss Transform (Yang et al., 2003)), **DFD** (dual-tree with finite-difference (Gray & Moore, 2003b)), **DFDO** (dual-tree with finite-difference and improved error control (Section 3.2)), **DFTO** (dual-tree with $O(p^D)$ expansion (Lee et al., 2006) and improved error control), and **DITO** (dual-tree with $O(D^p)$ expansion and improved error control).

All times (which include preprocessing but exclude parameter selection time) are in CPU seconds on a dual Intel Xeon 3 GHz with 2 Gb of main memory/1 Mb of CPU cache. Codes are in C/C++, compiled under $-\mathbf{O6}$ $-\mathbf{funroll-loops}$ flags on Linux kernel 2.6.9-11. The measurements in columns two to eight are obtained by running the algorithms at the bandwidth $kh^*$ where $10^{-3} \leq k \leq 10^3$ is the constant in the corresponding column. The dual-tree algorithms all achieve the error tolerance automatically. We also note that the **FGT** uses a different error tolerance definition: $|\widetilde{G}(x_q) - G(x_q)| \leq W\tau$. We first set $\tau = \epsilon$, halving it until the error tolerance $\epsilon$ was met. For the **IFGT**, we created an automatic scheme to tweak its multiple parameters based on recommendations given in the paper and software documentation: For $D = 2$, use $p = 8$; for $D = 3$, use $p = 6$; set $\rho_x = 2.5$; start with $K = \sqrt{N}$ and double $K$ until the error tolerance is met. When this failed to meet the tolerance, we resorted to additional trial and error by hand. We are primarily concerned with the sum of the times over all the bandwidths, shown in the last column of the table. Entries in the tables of 'X' denote cases where the algorithm exhausted RAM and caused a segmentation fault. Entries of $\infty$ denote cases where no setting of the algorithm's parameters was able to satisfy the error tolerance.

```
DITOBase(Q, R)
  forall x_q ∈ Q
    forall x_r ∈ R
      c = K_h(||x_q − x_r||), G_q^{min} += c,
      G_q^{max} += c, G_q^{est} += c
    G_q^{max} −= w_r
    Q.W_T += W_R, G_Q^{min} = min_{q∈Q} G_q^{min},
    G_Q^{max} = max_{q∈Q} G_q^{max}

DITO(Q, R)
  dl = W_R K_h(δ_{QR}^{max}), du = W_R K_h(δ_{QR}^{min}) − W_R
  W_T = W_R ( W|K_h(δ_{QR}^{min}) − K_h(δ_{QR}^{max})| / (2εG_Q^{min}) − 1 )
  // Optimized finite difference pruning first,
  if W_T ≤ 0 or Q.W_T ≥ W_T
    Q.W_T += −W_T, G_Q^{min} += dl, G_Q^{max} += du,
    G_Q^{est} += 0.5(dl + du + W_R), return
  else // FMM-type pruning
    {A, p, E_A} =
      bestMethod(Q, R, ε(W_R + Q.W_T)/W · G_Q^{min})
    if A = DH
      forall x_q ∈ Q
        G_q^{est} += EVALM(R.mcoeffs, p, x_R, x_q)
    else if A = DL
      Q.lcoeffs += DIRECTL(∀x_r ∈ R, p, x_Q)
    else if A = H2L
      Q.lcoeffs += H2L(R.mcoeffs, p, x_R, x_Q)
    if A ≠ DIRECT
      W_T = W_R ( WE_A / (εG_Q^{min}) − 1 ) − Q.W_T,
      Q.W_T = −W_T, G_Q^{min} += dl, G_Q^{max} += du,
      return
  if leaf(Q) and leaf(R), DITOBase(Q, R)
  else
    DITO(Q.l, R.l), DITO(Q.l, R.r)
    DITO(Q.r, R.l), DITO(Q.r, R.r)
```

Figure 7: The main procedure implementing a new error-control and $O(D^p)$ expansion.

```
DITOPost(Q)
  if leaf(Q)
    forall x_q ∈ Q
      G_q^{est} =
        EVALL(Q.lcoeffs, Q.p_L, Q.x_Q, x_q) + G_Q^{est}
  else
    G_{Q.l}^{est} += G_Q^{est}, G_{Q.r}^{est} += G_Q^{est}
    Q.l.lcoeffs += L2L(Q.lcoeffs, Q.p_L, Q.x_Q, Q.l.x_Q)
    Q.r.lcoeffs += L2L(Q.lcoeffs, Q.p_L, Q.x_Q, Q.r.x_Q)
    DITOPost(Q.l), DITOPost(Q.r)
```

Figure 8: Combining different types of approximations on different scales, using a breadth-traversal.

Our results demonstrate that the $O(D^p)$ expansion helps reduce the computational time on datasets of dimensionality up to 5. For example, on the 2-D dataset, the new algorithm **DITO** performed about 12 times as fast as the original **DFD** algorithm, which is in it-self an improvement over the naive algorithm. The datasets above five dimensions, however, present difficulty for the series expansion idea to be effective, and the new algorithm is slower than **DFD** algorithm. Yet the algorithm with the optimized pruning rule (**DFDO**) consistenyl yields about 10 % to 15 % improvement over **DFD** algorithm in higher dimensions.

| $sj2 - 50000 - 2, D = 2, N = 50000, h^* = 0.00139506$ | | | | | | | | |
|---|---|---|---|---|---|---|---|---|
| $Alg \backslash h^*$ | $10^{-3}$ | $10^{-2}$ | $10^{-1}$ | 1 | $10^1$ | $10^2$ | $10^3$ | $\Sigma$ |
| Naive | 452 | 452 | 452 | 452 | 452 | 452 | 452 | **3164** |
| FGT | X | X | X | 4.36 | 1.66 | 0.26 | 0.13 | **X** |
| IFGT | $\infty$ | $\infty$ | $\infty$ | $\infty$ | $\infty$ | $\infty$ | 7.05 | $\infty$ |
| DFD | 1.98 | 3.12 | 2.2 | 8.12 | 85.6 | 230 | 1.99 | **333** |
| DFDO | 2.02 | 3.18 | 2.19 | 7.08 | 77.7 | 170 | 0.82 | **263** |
| DFTO | 2.05 | 3.22 | 2.27 | 7.44 | 5.37 | 2.49 | 0.72 | **23.6** |
| DITO | 2.61 | 3.88 | 3.00 | 9.21 | 7.64 | 1.51 | 0.84 | **28.7** |

| $mockgalaxy - D - 1M - rnd, D = 3, N = 50000, h^* = 0.000768201$ | | | | | | | | |
|---|---|---|---|---|---|---|---|---|
| $Alg \backslash h^*$ | $10^{-3}$ | $10^{-2}$ | $10^{-1}$ | 1 | $10^1$ | $10^2$ | $10^3$ | $\Sigma$ |
| Naive | 461 | 461 | 461 | 461 | 461 | 461 | 461 | **3227** |
| FGT | X | X | X | X | $\infty$ | $\infty$ | $\infty$ | **X** |
| IFGT | $\infty$ | $\infty$ | $\infty$ | $\infty$ | $\infty$ | $\infty$ | $\infty$ | $\infty$ |
| DFD | 1.37 | 1.40 | 1.32 | 0.96 | 1.29 | 57.6 | 552 | **616** |
| DFDO | 1.40 | 1.43 | 1.35 | 0.97 | 1.25 | 44.5 | 355 | **406** |
| DFTO | 1.45 | 1.48 | 1.41 | 1.03 | 1.37 | 20 | 28.3 | **55** |
| DITO | 2.29 | 2.32 | 2.28 | 1.92 | 2.28 | 40.6 | 8.65 | **60.3** |

| $bio5 - rnd, D = 5, N = 50000, h^* = 0.000567161$ | | | | | | | | |
|---|---|---|---|---|---|---|---|---|
| $Alg \backslash h^*$ | $10^{-3}$ | $10^{-2}$ | $10^{-1}$ | 1 | $10^1$ | $10^2$ | $10^3$ | $\Sigma$ |
| Naive | 491 | 491 | 491 | 491 | 491 | 491 | 491 | **3437** |
| FGT | X | X | X | X | X | X | X | **X** |
| IFGT | $\infty$ | $\infty$ | $\infty$ | $\infty$ | $\infty$ | $\infty$ | $\infty$ | $\infty$ |
| DFD | 5.59 | 6.49 | 13.5 | 17.1 | 128 | 577 | 169 | **917** |
| DFDO | 5.75 | 6.67 | 13.7 | 16.2 | 113 | 544 | 81.6 | **781** |
| DFTO | 5.80 | 6.70 | 13.8 | 16.5 | 123 | 422 | 282 | **870** |
| DITO | 6.92 | 7.86 | 15.6 | 19.3 | 133 | 365 | 6.10 | **554** |

| $pall7 - rnd, D = 7, N = 50000, h^* = 0.00131865$ | | | | | | | | |
|---|---|---|---|---|---|---|---|---|
| $Alg \backslash h^*$ | $10^{-3}$ | $10^{-2}$ | $10^{-1}$ | 1 | $10^1$ | $10^2$ | $10^3$ | $\Sigma$ |
| Naive | 511 | 511 | 511 | 511 | 511 | 511 | 511 | **3577** |
| FGT | X | X | X | X | X | X | X | **X** |
| IFGT | $\infty$ | $\infty$ | $\infty$ | $\infty$ | $\infty$ | $\infty$ | $\infty$ | $\infty$ |
| DFD | 14.9 | 15.1 | 16.6 | 37.7 | 50.8 | 372 | 625 | **1132** |
| DFDO | 15.5 | 15.6 | 17.3 | 38.2 | 49 | 321 | 587 | **1044** |
| DFTO | 15.6 | 15.6 | 17.4 | 38.4 | 50.2 | 337 | 621 | **1095** |
| DITO | 16.5 | 16.7 | 18.4 | 40.5 | 54.7 | 362 | 703 | **1212** |

| $covtype - rnd, D = 10, N = 50000, h^* = 0.0154758$ | | | | | | | | |
|---|---|---|---|---|---|---|---|---|
| $Alg \backslash h^*$ | $10^{-3}$ | $10^{-2}$ | $10^{-1}$ | 1 | $10^1$ | $10^2$ | $10^3$ | $\Sigma$ |
| Naive | 515 | 515 | 515 | 515 | 515 | 515 | 515 | **3605** |
| FGT | X | X | X | X | X | X | X | **X** |
| IFGT | $\infty$ | $\infty$ | $\infty$ | $\infty$ | $\infty$ | $\infty$ | $\infty$ | $\infty$ |
| DFD | 26.5 | 29.7 | 88.2 | 104 | 557 | 659 | 11.4 | **1476** |
| DFDO | 27.2 | 30.5 | 90.2 | 98.2 | 515 | 623 | 5.73 | **1390** |
| DFTO | 27.4 | 30.7 | 90.6 | 101 | 477 | 660 | 6.10 | **1393** |
| DITO | 28.4 | 31.6 | 92.8 | 106 | 490 | 668 | 6.19 | **1423** |

| $CoocTexture - rnd, D = 16, N = 50000, h^* = 0.0263958$ | | | | | | | | |
|---|---|---|---|---|---|---|---|---|
| $Alg \backslash h^*$ | $10^{-3}$ | $10^{-2}$ | $10^{-1}$ | 1 | $10^1$ | $10^2$ | $10^3$ | $\Sigma$ |
| Naive | 558 | 558 | 558 | 558 | 558 | 558 | 558 | **3906** |
| FGT | X | X | X | X | X | X | X | **X** |
| IFGT | $\infty$ | $\infty$ | $\infty$ | $\infty$ | $\infty$ | $\infty$ | $\infty$ | $\infty$ |
| DFD | 19.3 | 36.6 | 107 | 199 | 611 | 641 | 0.56 | **1614** |
| DFDO | 19.9 | 36.4 | 107 | 237 | 589 | 375 | 0.58 | **1365** |
| DFTO | 20.1 | 37.8 | 108 | 189 | 629 | 401 | 0.60 | **1386** |
| DITO | 26.2 | 38.9 | 112 | 196 | 655 | 437 | 0.62 | **1466** |